# Whither Bias Goes, I Will Go: An Integrative, Systematic Review of Algorithmic Bias Mitigation

*Integrative Conceptual Review*

Louis Hickman[1], Christopher Huynh[1*], Jessica Gass[1*], Brandon Booth[2], Jason Kuruzovich[3], & Louis Tay[4]

[*]The second and third authors contributed equally.

[1]Virginia Tech

[2]University of Colorado-Boulder

[3]Rensselaer Polytechnic Institute

[4]Purdue University

**Author note:** Portions of this manuscript were presented at the 2024 Society for Industrial and Organizational Psychology Annual Conference. Thank you to Pat Dunlop, Andrew Speer, and Caleb Rottman for their feedback on earlier versions of the manuscript.

# Whither Bias Goes, I Will Go:

## An Integrative, Systematic Review of Algorithmic Bias Mitigation

*Abstract*. Machine learning (ML) models are increasingly used for personnel assessment and selection (e.g., resume screeners, automatically scored interviews). However, concerns have been raised throughout society that ML assessments may be biased and perpetuate or exacerbate inequality. Although organizational researchers have begun investigating ML assessments from traditional psychometric and legal perspectives, there is a need to understand, clarify, and integrate fairness operationalizations and algorithmic bias mitigation methods from the computer science, data science, and organizational research literatures. We present a four-stage model of developing ML assessments and applying bias mitigation methods, including 1) generating the training data, 2) training the model, 3) testing the model, and 4) deploying the model. When introducing the four-stage model, we describe potential sources of bias and unfairness at each stage. Then, we systematically review definitions and operationalizations of algorithmic bias, legal requirements governing personnel selection from the United States and Europe, and research on algorithmic bias mitigation across multiple domains and integrate these findings into our framework. Our review provides insights for both research and practice by elucidating possible mechanisms of algorithmic bias while identifying which bias mitigation methods are legal *and* effective. This integrative framework also reveals gaps in the knowledge of algorithmic bias mitigation that should be addressed by future collaborative research between organizational researchers, computer scientists, and data scientists. We provide recommendations for developing and deploying ML assessments, as well as recommendations for future research into algorithmic bias and fairness.

**Whither Bias Goes, I Will Go:**

**An Integrative, Systematic Review of Algorithmic Bias Mitigation**

Machine learning (ML) models are increasingly used for decision-making throughout society, including for recruitment and selection. ML is used to identify prospective applicants, screen resumes (Feffer, 2016), and score interviews (Hickman et al., 2022) and other open-ended assessments (e.g., essays, assessment center exercises; Campion et al., 2016; Hickman et al., 2023; Thompson et al., 2023). To create ML models, ML algorithms are applied to learn patterns (i.e., "training") in data. Trained ML models can also help with understanding and predicting complex behaviors such as career choice (Song et al., 2023) and turnover (Somers, 1999).

Vendors of ML pre-hire assessments (i.e., ML models for evaluating job candidates) claim that these tools enhance fairness and reduce bias compared to human decision-making (Raghavan et al., 2020). However, legislators, researchers, and consumer advocacy groups are concerned that ML models may perpetuate or exacerbate inequalities (Harris et al., 2019; EPIC, 2019; Landers & Behrend, 2022; Tippins et al., 2021). Given the problem's interdisciplinary nature, these concerns are complex as they span two distinct but often confounded issues of fairness and bias—along with a cornucopia of fairness and bias terms used by data scientists—suggesting a need to integrate psychometric, technical, and legal perspectives on these issues.

Industrial-organizational psychologists have begun to address these concerns, focusing on domain-specific approaches despite the interdisciplinary nature of algorithmic bias mitigation. Psychology has addressed ML audits (Landers & Behrend, 2022) and ML measurement bias (Tay et al., 2022), focusing on understanding legal and *psychometric* approaches. However, the field has not integrated the many technical solutions for mitigating algorithmic bias proposed by computer scientists (e.g., Bellamy et al., 2019), along with their specific terminologies for fairness operationalizations, which is needed to address algorithmic bias in personnel selection. Conversely, past technical solutions have not explicitly focused on whether these solutions comply with legislation outlawing disparate treatment, subgroup norming, and disparate (or *adverse*) impact in hiring (Civil Rights Acts of 1964 & 1991). Further, although many definitions and operationalizations of algorithmic fairness and bias have also been proposed (Mehrabi et al., 2022; Verma & Rubin, 2018), it is unclear how these fairness operationalizations (mis)align with psychometric notions of bias (Landers & Behrend, 2022) and adverse impact.

Considering these gaps and the rapid adoption of ML models, there is a pressing need to understand, clarify, and integrate multiple perspectives on algorithmic bias. This review aims to advance our understanding of technical, algorithmic bias mitigation approaches from computer science in a non-technical manner. Specifically, we map the proposed algorithmic bias mitigation solutions onto the ML development life cycle and detail their effectiveness, relationship to fairness operationalizations and psychometric bias, and legality for personnel selection.

To achieve this, Figure 1 presents a four-stage model of the ML development life-cycle, beginning with (1) generating the training data, (2) training the ML model, (3) testing the ML model, and (4) deploying the ML model for high-stakes assessment. We conduct a systematic literature review to uncover and integrate fairness operationalizations and algorithmic bias mitigation methods. At each stage of our model, we integrate psychometrics and computer science fairness operationalizations with legal requirements governing personnel selection in the United States and Europe, then review algorithmic bias mitigation methods through these lenses.

In doing so, our review contributes to industrial-organizational psychology in several ways. First, much work on ML assessments is being conducted in other fields, and organizational scholars must heed these developments to maintain relevance in these cross-disciplinary domains (Landers & Behrend, 2022). Second, by integrating multiple perspectives on operationalizing fairness, we clarify a growing jingle-jangle problem where multiple terms may refer to the same concept or one term may refer to multiple concepts. Third, our review advances both research and practice by clarifying how each algorithmic bias mitigation method aims to alleviate bias and whether practitioners can or should consider using it. Overall, we hope this review improves our understanding of when and why algorithmic bias occurs and how to mitigate it.

## Machine Learning Models in Industrial-Organizational Psychology

Table 1 defines key terms relevant to ML and algorithmic bias. In general, ML for personnel assessment relies on supervised ML, where algorithms are trained to use predictors to model some outcome (e.g., job performance, dependability; often referred to as the "ground truth"), which "supervises" predictor use. The resulting models replicate patterns in their inputs. Any empirical predictor weighting scheme keyed based on observed outcomes—such as those used for biodata (e.g., Cucina et al., 2012) or to weigh multiple assessments together into a composite (e.g., OLS regression; Wainer, 1976)—can be considered supervised ML models. Table S1 in the additional online material provide additional detail on unsupervised and reinforcement learning, but we focus on supervised ML because organizations are rapidly adopting supervised ML (hereafter ML) due to time and cost savings (Campion et al., 2016). The promise of ML models is that they can assess more individuals quicker than humans in a consistent, replicable manner treating all applicants equally. Despite this promise, concerns exist throughout society about algorithmic bias in high-stakes decision-making contexts.

**Algorithmic Bias and Fairness**

Industrial-organizational psychology has long investigated bias and fairness, yet most computer science research on algorithmic bias mitigation has ignored this body of research.

Psychometric bias is systematic error in test scores that differentially affects test takers as a function of group membership (AERA, APA, & NCME, 2014; SIOP, 2018). This systematic error is generally caused by failing to include relevant information (construct deficiency) or including irrelevant information (construct contamination). *Measurement bias* occurs when this error systematically causes higher/lower scores for individuals from one or more subgroups, and *predictive bias* occurs when it causes systematic prediction errors for individuals from a specific subgroup. Figure 2 illustrates the relationships between psychometric bias and other concepts in our review. Although it is rare for psychometric bias to be invoked in computer science research, our review reveals that many fairness operationalizations align with these definitions.

Fairness is a broader social concept, and one lens related to fairness comes from a justice perspective (Colquitt et al., 2022). Distributive justice regards whether decisions follow allocation norms, such as equity, equality, and need. For example, one perspective on distributive justice is requiring equal group outcomes (SIOP, 2018). Procedural justice considers the consistency and unbiasedness of decisions as well as one's ability to challenge the decisions. For example, applicants may consider a hiring process procedurally unfair if it does not predict outcomes equally well across groups (i.e., predictive bias), if they did not have comparable access to constructs due to cultural differences, or if they are not equitably treated throughout the process (SIOP, 2018). Interpersonal justice regards perceptions of the respectfulness of communications about the decision, and informational justice regards whether such communications are perceived as explanatory and truthful. Although such perceptions of justice are important for understanding applicant attraction and reactions to AI assessments (Landers & Behrend, 2022; Langer & Landers, 2021), they fall outside the scope of our review. Our focus is on formal definitions of algorithmic bias (i.e., fairness operationalizations) and methods for reducing such bias. Notably, these various perspectives are related, as Figure 2 illustrates.

Many organizational psychologists likely expect *algorithmic bias* to refer only to *psychometric* bias, but in other fields, bias is often used to refer to mean group differences (e.g., Bellamy et al., 2019). Many technical approaches labeled as "debiasing" ML algorithms focus on eliminating group differences. Thus, we refer to operational definitions of algorithmic bias as fairness operationalizations, given that they may focus on group differences.

The *Standards* and *Principles* reject defining fairness as equal group outcomes because measures may capture actual group differences (i.e., the aggregate qualifications of people from different protected groups differ in the applicant pool). Thus, "outcome differences … do not indicate bias" but should trigger heightened scrutiny for psychometric bias (*Principles*, 2018, p. 38). Nevertheless, group differences are practically important because substantial adverse impact constitutes *prima facie* evidence of employment discrimination in the United States (*Uniform Guidelines*, 1978). Substantial adverse impact occurs when members of two demographic groups are hired at different rates, as indicated by statistical significance testing (Morris, 2016) or violating the four-fifths rule (Equal Employment Opportunity Commission, 1978).

When adverse impact occurs, organizations can defend against discrimination claims by demonstrating that the selection procedure is job relevant (i.e., valid) and justified by business necessity (Civil Rights Act, 1964). However, even when the adverse impact is due to actual differences, organizations often seek to mitigate adverse impact for ethical reasons, to reduce the chances of litigation, and to enhance organizational diversity (Oswald et al., 2016). Thus, our review focuses on methods proposed at any stage of the ML lifecycle for mitigating algorithmic bias, which encompasses psychometric bias, adverse impact, and other concepts (Figure 2).

**Machine Learning Algorithm Life Cycle**

To understand algorithmic bias and methods for mitigating it, it is crucial to consider the process of developing and deploying supervised ML algorithms. The four major steps in this process include: 1) generating training data, 2) training the model, 3) testing the model, and 4) deploying the model. We present these as sequential steps, but they are often completed in an iterative fashion. For example, if the model scores exhibit poor validity in testing, the organization may collect additional data and repeat the process until validity improves. Similarly, after deployment, the algorithm's psychometric properties may change over time (SIOP, 2018), eventually necessitating updated training data and model retraining.

Further, actions such as creating rigorous documentation of the data involved and decisions made throughout the development process can be used to judge the resulting model's fairness (Landers & Behrend, 2022). Detailed documentation enables auditing the training data and process for potential concerns, which can then spur remediation (Bandy, 2021). Putting each major decision into writing could cause ML model developers to think more deeply about their decisions and the resulting effects (e.g., model cards; Mitchell et al., 2019).

Table 2 provides an overview of the four major steps, which we detail below. Although some concerns are relevant to multiple stages, we describe them where they first become relevant in the ML model lifecycle. Table 3 summarizes some potential sources of algorithmic bias relevant to personnel selection, including several from the data science perspective.[1]

### Generating Training Data

Generating training data is the process of designing and collecting data and extracting quantitative information from unstructured data (e.g., feature engineering). Important considerations at this stage include the sample characteristics (including demographics), choice of predictors and "ground truth" dependent variable, and the properties of the measures employed (Landers & Behrend, 2022). A primary concern regarding sample characteristics is the representation of different demographic subgroups in the training data, both in terms of sample sizes and mean scores. Supervised ML tends to capture the most prevalent patterns in training data. Thus, imbalanced subgroup representation in training data—one form of *representation bias*—is concerning if predictor-outcome relationships differ across subgroups. For example, if predictor-outcome relationships differ across genders, and the training data consists primarily of men, then the fitted model will tend to be more accurate for men (Barocas & Selbst, 2016).

---

[1] In these approaches, *bias* is often used synonymously with *fairness*, given that psychometrics defines bias as error in measurement or prediction that differs across groups (Tay et al., 2022).

The concept of *historical bias* suggests it is unfair for a ML model to accurately reflect the world if doing so will harm a subgroup (e.g., through uneven hiring rates; Mehrabi et al., 2022; also see Aguinis & Culpepper, 2024, for related arguments). Such group mean differences in training data may be caused by factors upstream of the focal ML model (e.g., unequal access to constructs) but lead the model to encode these disparities. Thus, researchers have suggested that we should consider not only the fairness of contests (i.e., fairness of algorithmic decisions) but also whether people had similar opportunities over their lifetime, which affects the data generation process (e.g., in the United States, children from low socioeconomic status (SES) households generally attend less well funded public schools; Khan et al., 2022). Further, the magnitude of subgroup mean differences on the training data ground truth relates linearly to the adverse impact of the ML model's predictions, raising concerns that training data with large ground truth mean differences can entrench or exacerbate inequality (Zhang et al., 2023).

The choice and properties of the predictors and "ground truth" variable encompass multiple concerns relevant to *measurement bias*. First, the relevance and coverage of predictors is important, as measurement bias may be caused when assessments are contaminated with construct-irrelevant variance or when they are deficient, by failing to capture construct-relevant variance (SIOP, 2018). For example, structured interviews focus on evaluating interviewee answers (Campion et al., 1996). Thus, when automating their assessment, using nonverbal or paraverbal behaviors for scoring would cause contamination, and failing to use verbal behaviors (i.e., the words spoken) would cause construct deficiency. Using theory to select meaningful predictors (i.e., content validation) and to derive them from naturalistic data (e.g., Sajjadiani et al., 2019) is likely a useful approach to addressing this concern (Van Iddekinge et al., 2023).

Second, both predictors and the ground truth may be imbued with *measurement bias*. For predictor variables, this could occur when a traditional assessment is biased. It can also occur in modern approaches because ML models that measure verbal, nonverbal, and paraverbal behaviors from natural language, video, and audio data may be biased (Hickman et al., 2022). For example, computerized transcription services used when automatically scoring interviews are less accurate for people with accents as well as Black and African American speakers compared to White speakers (Hickman et al., 2024; Koenecke et al., 2020). Further, when natural language processing is used to convert natural language text to numbers, word embeddings may exhibit differential, stereotypic associations for male and female pronouns (Bolukbasi et al., 2016), potentially resulting in bias in the model predictions (Tay et al., 2022).

Further, the choice and measurement of the ground truth variable holds the same potential for bias. For example, research with the COMPAS criminal recidivism dataset has sometimes used whether defendants were released on bail as the ground truth (Bao et al., 2021). However, this ground truth is a proxy variable for the actual phenomenon of interest (i.e., recidivism) that is contaminated with irrelevant information, causing measurement bias. Being released on bail is contaminated with SES, because low SES individuals may be unable to afford bail. Thus, selecting meaningful ground truth variables that best reflect the intended construct is critical

Regarding the measurement of the ground truth, *human bias* occurs when ML models rely on biased human ratings. Human bias comes in two forms: general and group-specific. General biases include biases like halo error that affect all assessees equally. Group-specific biases differentially affect assessees based on their characteristics. For example, the ground truth for training models to automatically score pre-employment assessments is often human ratings. However, assessor evaluations may be biased by irrelevant factors such as applicant race (Quillian et al., 2017), gender (Schaerer et al., 2023), attractiveness (Hosoda et al., 2003), and accents (Spence et al., 2024).

### *Training the Model*

Training (or *fitting*) a supervised model involves mathematically estimating relationships between predictors and the outcome variable. This results in a model with fitted parameters that can score future, unseen cases. The decisions relevant to fairness at this step include the type of model(s) to train and the model's optimization (or *loss*) function (Landers & Behrend, 2022).

The type of algorithm (e.g., random forest, ordinary least squares [OLS] regression, neural network) used for training is relevant because a given model may underfit or overfit the training data, which relates to the bias-variance tradeoff (in this tradeoff, "bias" refers to error from mistaken assumptions made by the algorithm; Yarkoni & Westfall, 2017). When a model overfits the training data, it minimizes prediction errors in the training data (i.e., low bias), but its parameters may change drastically when trained on another random sample drawn from the same population (i.e., high variance), resulting in poor generalizability and cross-validity. When a model underfits the training data, it exhibits greater prediction errors in the training data (i.e., high bias) but is less likely to change drastically when trained on a new sample (i.e., low variance). The resulting model may not accurately capture the predictor-outcome relationships.

In data where all predictor-outcome relationships are linear, OLS regression is the best estimator for the training data. Yet if the training data contains sampling error, the fitted model will inherit this error, be overfitted, and perform poorly when applied to new samples. Thus, many ML methods include regularization terms that help address this by minimizing the risk of overfitting to the training data (Putka et al., 2018). However, any linear model will be underfitted to the training data when the predictor-outcome relationships include curvilinear effects and interactions (i.e., high bias). When such effects are present, using more complex models that implicitly handle nonlinear effects, such as random forests, support vector machines, and neural networks, can reduce underfitting. Underfitting or overfitting can cause algorithmic bias by resulting in models that are, respectively, deficient or contaminated.

Relatedly, *aggregation bias* occurs when one predictive model is applied across multiple groups, even though the predictor-outcome relationships differ between groups (Suresh & Guttag, 2021). Psychological mechanisms may cause predictor-outcome relationships to differ between groups in selection (Aguinis & Culpepper, 2024) and elsewhere (e.g., David et al., 2023). In such a situations, applying one model for all will be suboptimal for both validity and fairness, even if—as is the case for selection in the United States—doing so is required by law.

Many modern prediction methods also include hyperparameters that affect model functioning. For example, the ridge, least absolute shrinkage and selection operator (LASSO), and elastic net regularization terms have a weight assigned to them that can be tuned to identify the optimal amount of regularization. Neural networks have many hyperparameters, including the learning rate, number of training epochs, number of hidden layers, and more. While it is well-known that hyperparameter choices affect model accuracy/validity, hyperparameter choices also influence algorithmic bias (Cruz et al., 2021; Liang et al., 2023). Thus, if algorithmic bias is ignored during hyperparameter tuning, this could exacerbate algorithmic bias.

Regarding the optimization/loss function, most predictive models' optimization functions focus on minimizing error upon fitting to the training data. For example, OLS regression identifies the parameters that minimize the sum of squared errors between the observed values and the fitted regression line. However, *learning bias*—or when the trained ML model exacerbates score differences between groups (Suresh & Guttag, 2021)—may occur. For example, an ML model may accurately recover ground truth group means but exhibit smaller variances than the ground truth scores (e.g., Fan et al., 2023), thereby exacerbating standardized group differences (Cohen's *d*). Thus, beyond adding regularization terms to improve cross-validity, fairness constraints can be added to ML model optimization functions (Rottman et al., 2023; Zhang et al., 2023). For example, multi-objective optimization research focuses on models trained to simultaneously optimize validity and minimize group differences (Song et al., 2017).

### Testing the Model

Testing the model involves estimating the psychometric properties of the trained model's scores in data not used during model hyperparameter tuning or training. Important considerations include the testing process, which psychometric properties to evaluate, the criteria set for determining adequate validity, and the characteristics of the test data relative to training and deployment data (Landers & Behrend, 2022). The testing process regards the type of cross-validation used. For example, cross-validation can be conducted with a single train/test split, *k*-folds, or in a temporally lagged sample (Landers & Behrend, 2022). A single train/test split from one sample is least rigorous because it is prone to sampling error, while temporally lagged cross-validation is most rigorous, because it reflects how the model will be used during deployment.

When evaluating the psychometric properties of ML scores, extensive guidance is provided by the Uniform Guidelines on Personnel Selection (1978), the *Standards for Educational and Psychological Testing* (2014), and the *Principles for the Validation and Use of Personnel Selection Procedures* (2018). We provide additional detail on the psychometric properties that should be investigated in the additional online material.

The characteristics of the test data are relevant in relation to both the training data and future data that will be encountered during deployment. When the test data is drawn from the same sample as the training data (e.g., as in single train/test splits or *k*-fold cross-validation), the ML scores' psychometric properties may be inflated compared to what will be observed during deployment (Putka et al., 2018). Regardless from where the test data is drawn, it should reflect the applicant population that will be encountered during deployment. When the test data and deployment population are mismatched (e.g., the training and test data come from job incumbents, but the model will be used with applicants), ML psychometric properties—including group differences—may be inaccurately estimated.

### *Deploying the Model*

Deploying the model occurs when the decision has been made to use a trained ML model for high-stakes assessment. Important considerations include how applicants react to the way algorithms are used in the assessment process, what information is shared with applicants (Landers & Behrend, 2022), consistency in the ML algorithm scores' psychometric properties compared to those observed during model testing (e.g., that there is minimal shrinkage; Song et al., 2017), and how the scores are used to make selection decisions. Additionally, many concerns from prior stages are relevant to deployment as well (e.g., measurement bias in predictors; similarity of training, test, and deployment populations), but we do not describe them again here.

Regarding applicant reactions, negative applicant reactions may increase the likelihood of litigation (McCarthy et al., 2017). Given that applicants tend to react negatively to algorithmic assessments (Langer & Landers, 2021), attending to applicant reactions and seeking ways to improve them is important for organizations. For example, automatically scored interviews are viewed less favorably if they use facial expressions for evaluation (Langer, Baum, et al., 2021).

When it comes to the information shared with applicants, many concerns arise. For example, proponents of explainable AI claim that sharing information about how the model was developed and its validity could improve reactions to AI (Langer, Oster, et al., 2021). At a minimum, applicants should be informed that they are being evaluated by ML.

*Emergent bias* (Köchling & Wehner, 2020) deserves note as an additional concern that arises during deployment. Emergent bias occurs when predictor-outcome relationships shift after model deployment, which can cause a previously unbiased model to become biased. This can occur for several reasons, including shifting relationships, use in a new population, or changes to the data generation process during deployment. For example, the widespread availability of large language models (LLMs) means that some work is now completed with the aid of LLMs and raises concerns that applicants may complete pre-employment assessments with LLMs (e.g., Canagasuriam & Lukacik, 2024).

Thus, after deployment, it is important to monitor the psychometric properties of ML model scores for changes, including adverse impact, measurement bias, predictive bias, and validity shrinkage. Psychometric properties may change upon deployment for several reasons, including due to differences in the test sample and deployment sample populations (either due to true differences or sampling error in the test sample), item leaks (i.e., validity may decrease if test information leaks online; Wainer, 2000), coaching effects, and because the true predictor-outcome relationships shift after deployment (e.g., from emergent bias).

Finally, how scores are used to make selection decisions may also affect the psychometric properties of algorithmic decision support systems. Decision-makers may frequently resist blindly following the recommendations of algorithms (Dietvorst et al., 2018). Thus, allowing them to modify the algorithm or its outputs can increase algorithm use, but it also tends to decrease the validity of decisions (Dietvorst et al., 2018; Downes et al., 2023).

Put together, there are many potential sources of algorithmic bias. Our integrative review seeks to uncover algorithmic fairness operationalizations and methods for addressing algorithmic bias. Further, we connect these notions to legal requirements in the United States and Europe.

## Method

**Transparency and Openness Statement**

We followed the *Journal of Applied Psychology* methodological checklist, and the data generated by our review is available on OSF together with additional online material: https://osf.io/8tp2j/. Our focus was on the classification, regression, and other metrics (i.e., fairness operationalizations), as well as the bias mitigation methods, their categorization, and descriptions. The study did not involve human subjects data so did not require IRB approval.

**Systematic Review**

We conducted a systematic literature review following the PRISMA guidelines. We used multiple search terms in computer science and organizational research databases, as displayed in Figure 3. We also used ancestor and child searches of core articles on the topic. We systematically screened the literature for two types of papers: (1) those that propose or review operational fairness definitions, and (2) those that propose or investigate algorithmic bias mitigation methods. Our initial search uncovered 3,321 potentially relevant articles. After the various screening stages, we sought 382 articles and retrieved 373 of them. During our review, we identified 51 additional articles that did not define any fairness operationalizations or bias mitigation methods. We additionally identified six articles through child and ancestor searches. Thus, our final review comprised 328 articles.

After the initial search, the first, second, and third authors screened the titles and abstracts. Then, the first author reviewed and coded all remaining articles, while the second and third authors each reviewed and coded about half of the remaining articles. Disagreements were resolved via discussion between the first three authors to generate a final set of article codes.

## Results

Figure 1 illustrates the connections among the ML model development life cycle, fairness operationalizations, legal requirements, and bias mitigation methods. Table 4 reports fairness operationalizations across the ML model lifecycle. Tables 5-9 report, respectively, for each segment of the life cycle: 1) conceptual categories of bias mitigation methods, 2) which fairness operationalizations are addressed by each category, and 3) whether the methods are likely legal according to United States anti-discrimination statutes and case law. In each section below, we discuss the sources of bias or fairness operationalizations addressed by each bias mitigation method and connect the fairness operationalizations to psychometric bias and adverse impact.

### Generating Training Data

#### *Fairness Operationalizations*

Our search uncovered several fairness operationalizations related to characteristics of the training data: *equivalence of computed features* (Tay et al., 2022), *adverse/disparate impact* (*Uniform Guidelines*, 1978), *training sample representativeness* (Tay et al, 2022), and *potential bias* (Minot et al., 2022). Equivalence of computed features concerns whether predictor variables have equivalent meaning across groups. Measurement bias could cause systematically higher/lower predictor scores or greater measurement error for members of certain subgroups.

Adverse impact regards group differences on the ground truth variable. Researchers consider ground truth group differences concerning because the trained ML model may encode these group differences (Barocas & Selbst, 2016). Related terms and metrics have also been used to operationalize the fairness of group differences in training data—such as the Gini coefficient, which is a measure of outcome inequality between groups (e.g., Siddique et al., 2020).

Training sample representativeness regards whether members of different groups are adequately represented (Sha et al., 2022) and has also been referred to as representation rate, data coverage, distribution bias, and imbalance ratio. When members of a group are underrepresented in the training data, representation bias may occur—particularly if their predictor-outcome relationships differ from members of other groups. Further, if members of one group are underrepresented and there is adverse impact favoring members of the overrepresented group, then the ML model may be less accurate at differentiating low and high scores for underrepresented group members, as compared to overrepresented group members.

Potential bias is the extent to which group membership can be predicted from the training data (Minot et al., 2022). If the predictors can predict group membership, it suggests that their use in the ML model could exacerbate group mean differences, regardless of the cause of those differences. We provide additional detail about potential bias in the additional online material.

### Legal Requirements

To our knowledge, no laws in the United States or anywhere else govern ML model training data characteristics. This means that training data is not required to be representative of the population to which the algorithm will be applied nor to include underrepresented groups .

Notably, if the relationships between predictors and outcomes are equivalent for people from all demographic groups, then representation bias is not a concern. That would align with traditional wisdom from industrial-organizational psychology, which says that predictive bias is rare in selection—particularly in terms of slope differences (e.g., Pfeifer & Sedlacek, 1971)—although more recent work has challenged this notion (Aguinis & Culpepper, 2024; Berry, 2015).

As described below, some preprocessing methods are applied during training and deployment. If transformations are applied during deployment, then the Civil Rights Acts (CRAs) of 1964 and 1991 apply to them in the United States. The CRA of 1964 outlaws disparate treatment based on protected characteristics, and the CRA of 1991 outlaws adjusting test scores (i.e., subgroup norming) based on protected characteristics. Similar statutes exist in the European Union and its member states outlawing direct (disparate treatment) and indirect (adverse impact) discrimination (e.g., EU Charter on Fundamental Rights, Article 21 – Non-Discrimination; Ellis, 2012). Thus, if a preprocessing approach requires access to demographic information during deployment, it is likely illegal because it constitutes disparate treatment and/or subgroup norming. However, when developing an ML model, the Equal Employment Opportunity Commission (EEOC) suggests making adjustments to reduce adverse impact (2023). No case law supports the EEOC's suggestion, but similar recommendations were provided by Rupp et al. (2020). Particularly, if an ML model with less adverse impact was considered but rejected during deployment, this could open the employer to liability.

### Preprocessing Bias Mitigation Methods

Bias mitigation approaches at this stage are known as *preprocessing* techniques since they modify the training data prior to model training. Our search uncovered 264 mentions of preprocessing methods. We uncovered three broad classes of preprocessing: (1) balancing representation, (2) removing demographic information from the predictors, and (3) removing demographic information from the ground truth.

Balancing representation can involve (a) sampling additional observations to balance training data sample sizes across groups (Taati et al., 2019), (b) oversampling (i.e., resampling and upsampling; Shahbazi et al., 2023; Zhang et al., 2023), (c) undersampling (i.e., downsampling; Castelnovo et al., 2020; Cheong et al., 2021), or (d) reweighing observations (Kamiran & Calders, 2012). Sampling additional observations involves deliberately collecting additional data from underrepresented groups (e.g., Taati et al., 2019). Oversampling involves sampling additional real or synthetic (e.g., via synthetic minority oversampling technique [SMOTE]; Chawla et al., 2002) observations to equalize group sample sizes, and undersampling involves dropping observations from groups with larger sample sizes (Castelnovo et al., 2020). Reweighing involves giving more weight to observations from underrepresented groups and/or less weight to observations from groups with larger sample sizes (Kamiran & Calders, 2012).[2] The hope is that addressing representation bias will result in models that better represent all groups' predictor-outcome patterns, reducing group disparities in error rates. Such approaches are legal, will exhibit the strongest effects when groups' predictor-outcome relationships differ, and perform least well when undersampling is used because it decreases training sample size.

Removing demographic data from predictor variables can involve (a) not including demographic information as predictors in models (i.e., suppression; Waller et al., 2023),[3] (b) removing predictors (Rottman et al., 2023), or (c) transforming predictors (e.g., Zemel et al., 2013). Excluding demographic information from ML models achieves fairness through unawareness and is a legal requirement for many demographic categories in the United States and elsewhere. While doing so does not guarantee that the ML model will be fair or unbiased, including demographic information as a predictor ensures differential functioning.

---

[2] Only some types of algorithms can accept observation weights, but oversampling and undersampling achieve similar results (Kamiran & Calders, 2012).
[3] Predictor selection approaches (e.g., removing predictors) could also be considered inprocessing techniques, yet they are decisions made prior to training the final model.

Removing predictors involves dropping predictors that exhibit large group differences. This method can be applied when training data is generated (a preprocessing approach, e.g., Verdugo et al., 2022) or during model training (a type of inprocessing technique, e.g., Rottman et al., 2023). The difference rests on whether the predictors are identified and removed prior to model training (preprocessing), or if the model is trained, then variables are removed based on their contribution to group differences in the ML model scores (inprocessing). The aim is to address proxy nondiscrimination, potential bias, and reduce predictor adverse impact. While this approach reduces group differences in ML model scores, it also tends to reduce validity.

Many predictor transformations have been proposed that aim to reduce or remove the relationship between demographics and predictor variables (e.g., learning fair representations, disparate impact remover; Bellamy et al., 2019; Zemel et al., 2013) and are among the most frequently studied algorithmic bias mitigation methods. These methods transform predictors to make them more equal across demographic groups in hopes of reducing ML model score group differences. However, the transformation applied to an individual differs depending on their demographics, and the transformation is applied during testing and deployment. These methods are *group-specific* (Rottman et al., 2023) because they apply different data transformations depending on an individual's demography. These transformations can mitigate algorithmic bias without major detriments to validity, but by being group-specific, they constitute disparate treatment—which is illegal for selection in the United States, Europe, and elsewhere.

Removing demographic information from the ground truth variable involves reducing group differences via oversampling or by massaging the ground truth (massaging is described in more detail in the additional online material). The same methods described above for balancing representation can be used to reduce, remove, or reverse ground truth group means and other distributional differences. Removing such differences addresses ground truth adverse impact and prevents group differences on predictor variables from being used to replicate ground truth group differences. Oversampling to reduce and reverse ground truth group differences is recommended because it does not require collecting new data, can be applied to any modeling scenario, and has only small effects on model accuracy (Zhang et al., 2023).

**Training the Model**

*Fairness Operationalizations*

Fairness operationalizations during model training regard *causal fairness*—or whether demographics affect ML model scores—and include *fairness through unawareness* (Hardt et al., 2016), *disparate treatment*, *counterfactual fairness*, and *fair inference* (Verma & Rubin, 2018). Fairness through unawareness (also known as anticlassification and fairness through blindness) considers ML models fair if they do not use demographic information during deployment.

Counterfactual fairness (Kusner et al., 2017) is similar to no unresolved discrimination, no proxy discrimination (Verma & Rubin, 2018), and total effect (Makhlouf et al., 2021). All regard whether demographic membership causally affects the predictors in the ML model—or whether the ML model scores depend on a descendent of a demographic variable. This involves using causal graphs (e.g., directed acyclic graphs) to illuminate how demographic variables affect predictor variables and how those predictor variables affect the outcome. Sometimes counterfactual fairness is examined by testing whether the ML model outcome remains the same if the individual were a member of a different demographic group, keeping all other predictors constant. However, this approach is too simplistic, given that outcomes will remain the same unless demographic membership is included as a predictor. Thus, counterfactual fairness involves not only modifying the demographic predictor(s) themselves but also all variables causally dependent on demography. However, comprehensively detailing such causal pathways is intractable, leading Verma and Rubin (2018) to suggest that "it is impossible to test" (p. 6) these causal fairness notions. For example, although there may be differences in how men and women use language, the potential differences are so numerous and culture-dependent that it becomes difficult to identify all the ways that gender affects language. These notions do, however, encourage scrutiny of possible predictors for their content relevance and potential bias.

Whereas most causal fairness notions only consider whether demography has a causal effect on ML model scores through predictor variables, fair inference considers whether those causal pathways are *legitimate* or *illegitimate* (Verma & Rubin, 2018). In the language of employment law, causal pathways are legitimate when predictors are justified by business necessity (Civil Rights Act of 1964), whereas illegitimate pathways have potentially been affected by discriminatory processes in the past. For example, credit scores are sometimes used in pre-employment testing, but historical biases affect them and they exhibit unclear relevance to job performance (Volpone et al., 2015). One way to address these concerns is to use predictors known to be meaningfully related to the criterion, regardless of adverse impact, and then considering additional predictors only if they do not increase adverse impact (Dutta et al., 2021).

### Legal Requirements

Although no United States laws explicitly govern ML model training, anti-discrimination employment (case) law (e.g., CRAs of 1964 and 1991) applies to model training. To avoid disparate treatment, ML models cannot use protected demographic information as predictors.

Including race or gender as a predictor would add or subtract points based on one's group membership, thereby constituting disparate treatment. While such bonus points were previously allowed in college admissions, the decision by the U.S. Supreme Court in *Students for Fair Admissions Inc. v. President and Fellows of Harvard College* (2023) similarly outlawed disparate treatment in college admissions. Further, because test scores cannot be adjusted or normed based on protected information, separate models cannot be developed for each subgroup in protected categories,[4] nor can scores be adjusted based on such information. Notably, the EEOC recommends and scholars agree that U.S. legal statutes and case law allow for using protected information during test development to reduce adverse impact (Rupp et al., 2020) and bias, although no case law has yet adjudicated these assertions. The *Uniform Guidelines* (1978) suggest, when possible, using assessment procedures with equal validity but smaller group differences. Fairness enhancing actions should be taken before using the selection system for high-stakes assessment, given that the courts consider post-hoc changes to avoid adverse impact in selection outcomes as a form of disparate treatment (*Ricci v. Destefano*, 2009).

Thus, protected information can be used before deployment to reduce adverse impact and/or psychometric bias. First, protected information could be used to develop a training sample that is likely to result in fairer ML model outcomes. Second, protected information could be used during model training (e.g., with multiobjective optimization) to develop a model that balances validity and a fairness operationalization (e.g., adverse impact, psychometric bias). Third, during model testing, protected information could be used to measure adverse impact and psychometric bias to choose among competing models. A key distinction between what is likely legal or not is whether protected demographic information is required during deployment—methods that require protected information during deployment are unlikely to be defensible because they constitute disparate treatment and/or subgroup norming.

---

[4] Some legal scholars argue that separate models for separate groups could be legal in the United States (Hellman, 2020), but we strongly disagree and argue that separate models constitute disparate treatment or subgroup norming.

*Inprocessing Bias Mitigation Methods*

During model training, bias mitigation strategies are called *inprocessing* techniques because they occur while the algorithm processes data during training. Our search uncovered 256 mentions of inprocessing methods. The vast majority of inprocessing methods modify the algorithm's loss function to simultaneously optimize validity *and* a fairness operationalization. A second approach trains separate models, and a third uses an initial ML model's predictions to reweigh observations (which we describe only in the additional online material). In industrial-organizational psychology, methods that simultaneously optimize validity and fairness are known as multi-objective optimization (e.g., Song et al., 2017). Most work in psychology focused on the normal boundary intersection (NBI) pareto optimization method implemented by De Corte et al. (2007), but Rottman et al. (2023) and Zhang et al. (2023) recently showed how to incorporate such constraints in other algorithms' loss functions (and this has been done in computer science too; e.g., adversarial learning; Zhang et al., 2018). Multi-objective optimization can maintain validity while minimizing adverse impact.

Further, work elsewhere has gone beyond adverse impact when incorporating such constraints, including minimizing a variety of distributional disparities or differences in ML model accuracy. The additional online material provides examples from our review. Such constraints could be combined with ones that minimize adverse impact to balance validity, group differences, and measurement bias (i.e., equal accuracy).

Notably, the NBI Pareto optimization approach seeks to identify equally spaced solutions between the two endpoints (i.e., where validity or diversity is maximized). However, because the predictor space can be much larger in algorithmic assessments, other approaches often try to find a single, optimized solution. In both cases, users can adjust the extent of diversity optimization—when using NBI, it is adjusted by the human choice of which model to use, and when using other algorithms, a hyperparameter usually determines the weight assigned to the fairness regularizer in the loss function. That hyperparameter can be tuned automatically (e.g., through cross-validation and setting a rule for which to choose) or through human judgment (i.e., by examining the possible results and choosing the one that strikes the desired balance between validity and fairness). The lack of attention to NBI in computer science suggests there is promise in these other approaches, especially when working with "big" data that involves hundreds or even thousands of predictors (e.g., Rottman et al., 2023; Zhang et al., 2023).

A less common approach considers multiple objectives during hyperparameter tuning. Many algorithms include tunable regularization terms. Hyperparameters often are tuned by identifying which provide the highest accuracy/validity in the training data, but maximizing only validity may harm ML model diversity. Thus, considering diversity and validity during hyperparameter tuning—or more broadly, when choosing between different algorithms (Black et al., 2022)—can improve diversity without sacrificing validity (Cruz et al., 2021).

A second class of methods has been proposed that aims to address aggregation bias and is much simpler to implement: training separate models for each group or including protected information as predictors (e.g., Kleinberg et al., 2018). Because such methods likely constitute subgroup norming and disparate treatment, they are likely illegal for personnel selection in the United States, Europe, and other countries that outlaw disparate treatment. However, such approaches can *improve* validity and reduce adverse impact (Kleinberg et al., 2018). Thus, using this approach in non-selection contexts may be useful (e.g., improving retention).

**Testing the Model**

*Fairness Operationalizations*

*Group fairness* during model testing regards group-level consistency in scores (i.e., *independence*) and validity (i.e., *separation* and *sufficiency*), including *adverse/disparate impact* (Uniform Guidelines, 1978), *conditional statistical parity*, *equal accuracy* (Berk et al., 2021), and *differential functioning* (Tay et al., 2022). Additionally, *individual fairness* regards *consistency* in scores assigned to individuals who are similar except for their demographics. Adverse/disparate impact is well-known to organizational researchers, given that the CRA of 1964 established that substantial adverse impact against members of a protected group constitutes *prima facie* evidence of discrimination. Computer scientists also refer to adverse impact as demographic/statistical parity and independence (i.e., ML model outcomes should be statistically independent of demographic membership). Other notions have gone further to consider all potential distributional differences, exemplified by the use of Jensen-Shannon and Kullback-Leibler divergence measures that capture differences in two sets of scores' distributions. All are concerned with potential score and outcome differences between groups.

Conditional statistical/demographic parity focuses on minimizing adverse impact conditioned on some legitimate factor(s) (Verma & Rubin, 2018). For example, if we know the job performance of a set of individuals, conditional statistical parity focuses on equal hiring rates across groups given equal levels of performance. This is similar to Thorndike's constant ratio and Cole's conditional probability definitions of fair employee selection, which say that selection is fair when subgroup selection ratios are proportional to their success ratios (Steffy & Ledvinky, 1989). These notions align with the *Standards*' rejection of fairness as a lack of group differences, given that actual mean group differences in job-relevant qualifications may occur.

Equal accuracy (Chakraborty, Peng et al., 2020) subsumes many labels and definitions because accuracy has been operationalized in many ways, but all regard group-level consistency in validity—or measurement bias and differential functioning—because they concern whether the ML scores are equally accurate/valid across groups. Adverse impact involves comparing outcomes between groups, while equal accuracy involves comparing a measure of ML model accuracy between groups. For continuous variables, equal accuracy can be operationalized by comparing the correlation between ground truth and ML scores across groups (Tay et al., 2022) or the amount of error (e.g., mean absolute error) across groups (Fu et al., 2021). For discrete (usually binary) variables, equal accuracy is operationalized by comparing elements of the confusion matrix across groups (e.g., true positive rates; false positive rates; Beutel et al., 2017). Such approaches, by far, represent the most common approach to operationalizing fairness. Confusion matrices (also known as classification tables) cross observed ground truth outcomes with ML predicted outcomes. A true (false) prediction is accurate (inaccurate), and a positive (negative) prediction is usually the desirable (undesirable) outcome (e.g., being hired).

Many fairness operationalizations for discrete variables (e.g., equal accuracy) are derived directly from a confusion matrix (e.g., adverse impact), including accuracy, which is the sum of true positive and true negative ML model predictions divided by the total number of predictions (i.e., true positives, true negatives, false positives, and false negatives). We summarize other common notions in the additional online material, and Makhlouf et al. (2021) provided a detailed summary of them. We recommend referring to the specific elements of the confusion matrix that are being compared (e.g., true positive, false positive, true negative, and false negative decisions) to avoid confusion and jingle-jangle concerns. Depending on which elements of the confusion matrix are being compared, different tradeoffs are being allowed. For example, focusing on equal true positive rates in selection implies that the organization values accurately identifying high performers across groups, while focusing on equal true negative rates implies the organization values not mistakenly rejecting people who might otherwise perform well across groups.

The notion of equal accuracy in computer science has focused primarily on whether the ML model is similarly accurate across groups in recovering the ground truth labels. However, *predictive bias* regards when prediction errors sum to zero across groups (Cleary, 1968), a form of differential functioning. When the ground truth variable is a predictor (e.g., grit) of the outcome of interest, then predictive bias involves relating ML scores to the outcome/criterion (e.g., job performance). When the ground truth variable *is* the outcome, then equal accuracy and predictive bias are equivalent

Recently, a standardized effect size metric for predictive bias was proposed: $d_{\text{Mod\_Signed}}$ (Nye & Sackett, 2017). The simplest formula for estimating this metric is: $d_{\text{Mod\_Signed}} \cong d_Y - r_{XY1}d_X$ (Dahlke & Sackett, 2022). This metric suggests that, when a selection procedure is unbiased, reducing group differences will cause predictive bias (unless $r_{XY1}$ or $d_Y$ is also changed). Thus, organizations may have to choose between a biased selection procedure or one that minimizes adverse impact. We suspect most organizations will choose to minimize adverse impact, given the lack of legal statutes regarding predictive bias.

Another approach to considering equal accuracy regards whether the variability of prediction errors is consistent across groups. For example, Xiang et al. (2022) proposed using the standard deviation of the mean absolute error of observations within groups. Such approaches recognize that, even when the mean error is the same across groups, it is concerning if one group has a subset of observations with excessively high error.

Consistency (also known as individual fairness and fairness through awareness; Dwork et al., 2012; Zemel et al., 2013) suggests that ML models are fair when an individual receives the same outcome as similar individuals. Defining similar individuals can be difficult, but most approaches rely on identifying *k* near neighbors and then examining the proportion of near neighbors who received the same outcome as the focal individual. This is particularly relevant when a focal individual is compared to the proximate individuals whose demography differs from theirs. The proportion of neighbors from a comparison demographic group who received the same outcome can be averaged across all observations to measure consistency.

One limitation of these fairness operationalizations is that real-world demographic data is often incomplete due to the voluntary nature of self-disclosure in the United States. Particularly when the data are not missing at random (e.g., members of a certain subgroup voluntarily self-identify less than members of a comparison subgroup), missing data may cause issues in the conclusions drawn from these comparisons. One solution is to use ML models to predict race and gender from names, images, and/or videos (e.g., Chekili & Hernandez, 2023; Rottman et al., 2023). However, since such models are imperfect, their predictions come with error which could also result in drawing incorrect conclusions. Further, adequate sample sizes are necessary for any such fairness operationalizations—for example, when sample sizes are small, adverse impact ratios can be drastically affected by changing just a few selection decisions.

*Legal Requirements*

The CRAs of 1964 and 1991 apply to how the models will be used. Because disparate treatment is outlawed, organizations must apply the same model regardless of protected demographics. Further, the ML model scores must be used consistently across legally protected subgroups because any changes to scores or score interpretation based on subgroup membership constitute disparate treatment and/or subgroup norming.

Regional legislation has emerged that is relevant to model testing. The New York City Local Law 144 requires that automated employment decision tools used for personnel selection in New York City must undergo an annual "bias audit" and publish a summary of the audit's results. The audit must calculate group means or selection ratios (i.e., adverse impact) across sex, race/ethnicity, and intersectional subgroups representing at least 2% of the data. Notably, labeling group differences as "bias" misaligns with the *Standards* and *Principles*, although group differences should trigger heightened scrutiny for psychometric bias. The audit can use test data or archival, high-stakes data. However, no action is required as a result of the audit.

Estimating adverse impact would also align with *The Uniform Guidelines on Selection Procedures* and potential liabilities raised by the CRA of 1964, given that adverse impact constitutes *prima facie* evidence of discrimination. An organization can defend against such evidence if it can demonstrate the job relevance of the assessment. Thus, at the stage of model testing, it is also desirable to establish multiple types of validity evidence, including content (i.e., assessment content matches job requirements, as identified by work analysis), relationships with other variables (i.e., convergent, discriminant, and criterion relationships), and lack of bias (i.e., measurement and predictive, given that bias can detract from the validity argument). The fairness operationalizations from the prior stages are relevant here as well, and each type of validity evidence can help an organization ensure that the algorithmic assessment is both fair and valid.

### Postprocessing Bias Mitigation Methods

These approaches are known as *postprocessing* techniques since they involve modifying model outputs (e.g., Kamiran et al., 2012). Our search uncovered 92 mentions of postprocessing methods. Although such techniques become part of the model ultimately used for scoring (thus, in a way, making them part of model training), they are not developed until after a model has been trained and undergone initial testing. Thus, we put them in the testing stage.

Specifically, after an ML model has been trained and then used to make predictions on test data, those predictions are evaluated. If the predictions violate a specified fairness operationalization, then a group-specific postprocessing method can be applied to the predictions to enforce some fairness operationalization—such as minimizing adverse impact or differences in true positive rates. These methods modify the outcomes of earlier stages of the process rather than addressing the cause(s) of algorithmic bias. For example, reject option classification (Kamiran et al., 2012) can be applied to the outcomes of a binary classifier, and it considers the predictions where the model is most uncertain (i.e., the predictions closest to the decision boundary). Then, reject option classification flips uncertain model predictions for disadvantaged (advantaged) group members from reject (accept) to accept (reject) to reduce adverse impact. Another common form of postprocessing, known as thresholding or calibration, involves setting separate thresholds across groups to achieve some fairness operationalization, such as demographic parity or equal error rates (e.g., Hardt et al., 2016; Pleiss et al., 2017).

However, much like score banding (i.e., treating scores within a certain range as equivalent to each other; Campion et al., 2001), these procedures only reliably reduce adverse impact if they use protected information to modify decisions (as in reject option classification) or set different thresholds (as in thresholding)—which would be illegal because it constitutes disparate treatment and/or subgroup norming.[5] Further, demographic information is often unavailable during deployment, and prior reviews suggest that postprocessing methods harm validity more than preprocessing or inprocessing methods (Zehlike et al., 2023). Overall, then, postprocessing methods are not likely useful or legal for personnel selection.

---

[5] Although postprocessing methods do exist that do not require demographic information at the time of inference (e.g., regressor distribution control; Miroshnikov et al., 2021), little empirical research examines their efficacy.

**Deploying the Model**

*Fairness Operationalizations*

Our search uncovered no fairness operationalizations specific to deployment that are not also relevant to earlier stages. Many operationalizations from earlier stages apply to deployment, including measurement bias in the predictor variables, causal fairness, group fairness, and individual fairness/consistency of the ML model scores. In other words, virtually all fairness operationalizations not concerned with the ground truth also apply to model deployment.

*Legal Requirements*

Any laws that apply to other selection procedures in the United States apply to deployed ML models, including the Civil Rights Acts of 1964 and 1991 and New York City Local Law 144. Thus, during deployment ML model scores should be monitored for adverse impact. Organizations should use work analysis to establish a link between assessment and job content, and collect multiple types of validity evidence. Additionally, group differences should trigger heightened scrutiny, particularly if those group differences are larger than observed during testing or if they increase over time. Doing so can help ensure validity and legality. Further, disparate treatment is outlawed, annual "bias" audits are required if operating in New York City, and Illinois requires informing job applicants if they will be evaluated by an ML model.

*Bias Mitigation Methods*

Because disparate treatment is outlawed, bias mitigation must occur prior to deployment. During deployment, protected demographic information cannot be used: as a predictor variable, to determine which transformation is applied to predictor variables, to determine which ML model will be applied to the individual, or to determine how to postprocess ML scores. Group agnostic approaches to score adjustments are acceptable—such as score banding. However, this may not reduce adverse impact unless demographic information is explicitly used for within-band decisions (Campion et al., 2001). Instead, organizations should continuously monitor the quality of algorithmic decisions and recommendations (Sackett et al., 2012), including examining ML model score group differences, the adverse impact of final decisions, validity in predicting work outcomes, and predictive bias (which such studies have not yet addressed).

To the extent that validity decreases, bias increases, or group differences increase over time, it may become necessary to retrain the model. For example, the literature on empirical biodata keying suggests that the keys should be re-validated every two to three years and re-keyed if validity has decreased (Reilly & Chao, 1982; Thayer, 1977). Retraining the model could necessitate collecting additional training data, adjusting the current training data, trying new types of models or mitigation methods, and/or using updated testing data.

## Discussion

Because work on algorithmic bias spans many fields, our work takes an important first step toward integrating the vast, varied perspectives on a topic that is central to industrial-organizational psychology. Integrating this literature for diverse audiences is important because some fairness operationalizations and algorithmic bias mitigation methods uncovered by our systematic review are likely outlawed by United States and European employment law. In many cases, there is substantial overlap across fields in how fairness is operationalized, considering the focus on adverse impact, differential functioning, and disparate treatment. However, there are also substantial gaps, including suggestions to use postprocessing methods that utilize demographic information to adjust scores (i.e., disparate treatment) and a lack of emphasis on predictive bias. Inprocessing approaches to multi-objective optimization have been shown to be effective across fields, and preprocessing approaches can provide additional benefits and are legal if they do not apply group-specific transformations during deployment.

**Methodological and Practical Implications**

Our review focused on works that specifically referred to algorithms or ML. Though we examined their application through the lens of personnel selection, ML methods are flexible and quickly becoming ubiquitous in society. However, an algorithm is, simply, a set of rules for converting inputs to outputs for a certain task. Thus, our review and its implications are potentially applicable to any approach for empirically deriving weights for any model intended for high-stakes decision making. Regardless of whether the context is biodata, traditional selection systems, social media, clinical psychology, criminal justice, or loan decisions, the same concerns, recommendations, and remedies apply.[6] The cross-disciplinary nature of these concerns suggests a need for additional cross-disciplinary reviews and research.

Although most bias mitigation methods focus on the context of two subgroups (e.g., White and Black applicants), it is possible to extend many solutions to handle multiple subgroup comparisons (Kozodoi et al., 2022). Indeed, preprocessing methods for balancing representation and ground truth means (Zhang et al., 2023), inprocessing multi-objective optimization techniques (Rottman et al., 2023), and postprocessing methods (Zehlike et al., 2023) can all be extended for multiple subgroup comparisons. However, increasing the number of subgroup comparisons increases the sample size needed to achieve a solution that will cross-validate with minimal shrinkage, so it should only be considered when sample sizes are adequately large. This limits the applicability of such approaches in cases of intersectional identities, given the rapidly increasing number of subgroups when intersectional identities are considered. Inprocessing approaches for reducing group differences in a single subgroup comparison require at least 300 observations to cross-validate with minimal shrinkage in traditional selection systems (Song et al., 2017), so multiple subgroup comparisons will require even more training data as well as adequate representation of each relevant subgroup.

---

[6] The legal requirements described in our review only apply to personnel selection. The Civil Rights Act of 1964 includes similar but distinct provisions for housing, employment, voter registration, and public education.

Multiple researchers have now noted that some fairness operationalizations cannot be simultaneously achieved in most real-world settings (e.g., Dahlke & Sackett, 2022; Kleinberg, 2018; Zhang et al., 2023). As noted above, if a predictor composite is unbiased, then changing the extent of group differences on the composite necessarily creates predictive bias—primarily in terms of intercept bias. Similar issues are observed when trying to balance elements of the confusion matrix across groups vs. minimizing adverse impact. Thus, as long as actual differences exist between demographic groups on predictors and outcomes, it will not be possible to simultaneously satisfy all fairness operationalizations. Vendors and organizations must decide which fairness operationalizations and perspectives are most important to them and their stakeholders, then seek to optimize and document those.

**Recommendations for Developing, Deploying, and Maintaining ML Models**

In addition to the effectiveness of the different approaches described above, our findings and information gleaned from the review have important implications for organizations. Table 9 reports our recommendations for developing, validating, and maintaining ML models. First, rigorous documentation of all decisions made and code used should be maintained. One example, prescriptive technique for documenting these decisions, any measurable group differences, and the context in which a model is meant to be used is called "model cards" (Mitchell et al., 2019). These cards capture preprocessing, inprocessing, and postprocessing decisions, metrics, intended uses, and ethical considerations and designer recommendations, but they are intended to serve as detailed summaries rather than comprehensive enumerations.

Approaches like this which also include all stakeholder inputs and decisions made throughout the ML process would enable organizations and independent auditors to examine the process and code for fairness concerns, as well as improve and update the models more easily—particularly if the person(s) responsible for developing the original model leave the organization. This is particularly important when considering adopting assessments from vendors, since the organization will likely be the one liable for any discriminatory outcomes. We have heard of vendors being unwilling to share such information, which is a major red flag.

When generating the training data, an effort should be made to collect diverse data that is representative of the intended population. In this context, diverse means diverse protected demographics, including gender, race, and more. Given that differences in predictor-outcome patterns across groups would cause bias if one group is overrepresented, diverse sampling can protect against this issue. The data should also be representative of the intended population, meaning that if it will be applied only to computer scientists, it should only include data from computer scientists—ideally matched in terms of the type of work they will be doing and the seniority level of the position. However, if the model will be applied to people in a variety of occupations, then the training sample should also include data from a variety of occupations.

Next, the fairness of the training data must be considered. First, adverse impact on the ground truth should be minimized. This can be accomplished through purposive sampling (i.e., sampling high-performing minorities), oversampling (e.g., Zhang et al., 2023), or reweighing. Second, all variables should be evaluated for measurement and predictive bias. Including biased predictors or ground truth in an ML model is likely to result in it being biased as well—if no steps are taken to minimize their influence on the final model (e.g., with inprocessing methods). Third, predictor variables with large group differences should be scrutinized for potential sources of bias (see Table 3) and job relevance. Predictors that cause adverse impact could open organizations to additional lawsuits and legal scrutiny.

When training the model, particularly if the model will be used for personnel selection, organizations should not use demographics as predictors in the model, to determine which model to apply to individuals, or to adjust their scores. Doing so constitutes disparate treatment. Thus, such approaches are unlikely to hold up to legal scrutiny in the United States or Europe.

Another important aspect of predictor choice is the meaningfulness and content relevance of predictors in the model. All variables' content relevance should be considered (SIOP, 2023). For traditional data, such as knowledge and skills, content relevance can be established via work analysis. However, this process is less straightforward for behavioral data and trace data, including text, video, and audio data. When using natural language text, the context of the language collection must be considered, rather than the meaningfulness of specific NLP predictors. Modern NLP embedding approaches do not generate inherently meaningful individual variables, but they excel in capturing the semantics of speech. So, organizations should seek to ensure that the text inputs are job relevant. For example, when automatically evaluating resumes, remove applicant names prior to evaluation.

Additionally, during model training, both validity and diversity should be optimized. If only model validity is optimized, diversity will suffer, and optimizing solely for diversity will harm validity and utility. Multi-objective optimization—including when comparing different algorithms/models, modifying loss functions (i.e., inprocessing methods), and during hyperparameter tuning—can frequently improve model diversity without substantial decrements to validity, making models that use multi-objective optimization preferable from a legal perspective (Uniform Guidelines, 1978).

When choosing a type of model (e.g., ridge regression vs. neural network), additional model complexity that reduces model interpretability must be justified by validity gains. More complex algorithms may increase the risk of issues such as bias since the complex relationships uncovered in the training data may not generalize to the population encountered during deployment. Further, it is difficult to elucidate and probe complex relationships in less interpretable models for bias. These risks should be weighed against any validity gains.

When testing the model, multiple aspects of validity evidence should be evaluated for the ML scores, in line with the *Standards* and *Principles*. Each piece of evidence helps build the validity claim. This includes examining the adverse impact, equal accuracy, and differential functioning of the ML model scores. Bias detracts from the validity argument, and if adverse impact is substantial, then alternative approaches should be explored to determine if less adverse impact can be achieved with similar levels of validity.

An important caveat here and when evaluating adverse impact for ML models is that multiple datasets are often combined for training and evaluation. These different subsets of data (e.g., different jobs) should be examined separately for validity and adverse impact during model testing. The risk is similar to Simpson's paradox, where phenomenon disappear or reverse depending on whether data are aggregated or analyzed as separate subsets. For example, contradictory adverse impact in subsets (e.g., adverse impact favoring men in one dataset but favoring women in another) could cancel out when the data are combined, such that only analyzing the combined dataset would suggest no adverse impact. However, an organization is responsible for the separate uses of an assessment in subsets of data, so psychometric investigations should be conducted at both the subset and overall levels.

Further, while *k*-fold cross-validation is commonly used, the ideal cross-validation approach involves a lagged sample collected after the model was trained. Given that this is the type of data the model will encounter during deployment, psychometric properties estimated this way should better reflect what to expect during deployment. This recommendation is similar to suggestions to examine multi-objective optimized predictor composites on pilot samples before deploying them for high-stakes assessment (e.g., Zhang et al., 2023).

Upon selecting and deploying a model for high-stakes assessment, the model's scores should routinely be re-evaluated. This involves examining multiple pieces of validity evidence, adverse impact, equal accuracy, and bias whenever possible. Empirically derived weights may eventually experience validity decrements (e.g., as in empirically keyed biodata), and changes in the population may result in emergent bias. In addition, shifts in the demographics, occupations, industries, or qualification-level of assessees could negatively affect the model scores' psychometric properties. Thus, these characteristics of individuals in the training and testing data should be compared against the high-stakes data. If decrements in validity occur, it may be beneficial to update the training data and retrain the model to recover the lost validity. However, organizations must weigh the cost of retraining against any potential utility gains.

**Future Directions**

Inprocessing bias mitigation methods are widely studied, but little work has examined how to effectively conduct multi-objective optimization for end-to-end deep learning with language models. Such models are rapidly being adopted in organizational research and practice (e.g., Speer et al., 2023; Thompson et al., 2023). Modern embedding methods are neural networks trained on masked language prediction, and they can be fine-tuned for other prediction tasks by adding a classifier or regressor layer to the neural network. We noticed very few articles in our review that conducted inprocessing with end-to-end deep learning with language models (for an exception, see: Attanasio et al., 2022). This shortcoming and the complexity of current inprocessing approaches for neural networks may explain why Rottman et al. (2023) modified ridge's loss function to minimize adverse impact in language-based ML models, as opposed to using end-to-end language modeling. Given the enhanced validity that can potentially be gained from end-to-end language modeling, future work should address this gap.

Relatedly, an emerging issue regards the potential use of LLMs such as OpenAI's Generative Pretrained Transformers (GPT) or Anthropic's CLAUDE for assessment and selection tasks. Zero-shot (i.e., no examples) or few-shot (i.e., a few training examples) prompts for LLMs can be used to evaluate the quality of open-ended assessment responses (Hickman & Liu, 2024). Further, LLMs can be used to query documents, such as resumes. However, we know little about the potential biases in these systems, and new approaches are needed for minimizing any biases, given that retraining them is prohibitively expensive. Initial research suggests diversity valuing prompts can mitigate LLM biases (Glazko et al., 2024).

To date, we know little about the effects of combining multiple debiasing methods. For example, oversampling to remove group differences in training data and using multi-objective optimization might, in combination, reduce group differences in the predictor composite more than using only one of the methods. Of the available research, ensembling—or using multiple types of preprocessing or inprocessing methods then averaging their results—worsens fairness operationalizations relative to using a single preprocessing or inprocessing method (Badran et al., 2023; Feffer et al., 2022). However, using multiple methods sequentially (e.g., preprocessing to remove ground truth group differences, then inprocessing multi-objective optimization) does improve upon using just one method (Feldman & Peake, 2021). However, additional research is needed to determine when combinations of methods are effective in high-stakes settings.

Some approaches to enhancing fairness that are not algorithmic bias mitigation methods deserve note. Participatory design regards involving stakeholders who will be affected by the use of AI in the design and development process (e.g., Trewin et al., 2019). This includes gathering input when generating the training data, selecting predictors, deciding on the algorithm's optimization function and which ground truth to use. Further, stakeholders can be involved in the testing process to evaluate if the trained model is likely to provide a net benefit to stakeholders.

Explainable AI (XAI) involves explaining the functioning of ML models or their specific outputs in a way that is understandable to humans (Langer, Oster, et al., 2021). Global XAI methods focus on reporting: the model's accuracy and validity, how the model was developed and what training data was used, cases that receive low or high scores from the trained model, and/or the model's most important predictors. Local XAI methods focus on how a particular case was scored and report influential predictors for the focal case's score, near neighbor example cases that would receive the same score, and/or near neighbor example cases that would receive a higher or lower score (Lai et al., 2023). Thus, XAI holds potential for uncovering fairness issues in the model development process as well as in the model's functioning itself.

Although our review focused on supervised ML, similar concerns exist for unsupervised ML and reinforcement learning. Given the empirical superiority of supervised ML for classification and regression problems (as in personnel assessment), most bias mitigation methods to date have focused on supervised ML. However, the characteristics of the training sample—including group mean differences on the predictors and ground truth variable—are concerns for any type of ML. To the extent that unsupervised ML and reinforcement learning grow in use for psychological, educational, and pre-employment testing, more research will be needed on robust methods for enhancing fairness and reducing bias in them.

ALGORITHMIC BIAS MITIGATION 41

ALGORITHMIC BIAS MITIGATION 44Electronic Privacy Information Center (EPIC). (2019). Complaint and request for investigation, injunction, and other relief. https://www.washingtonpost.com/context/epic-s-ftc-complaint-about-hirevue/9797b738-e36a-4b7a-8936-667cf8748907/

Equal Opportunity Employment Commission. (2023). *Select issues: Assessing adverse impact in software, algorithms, and artificial intelligence used in employment selection procedures under Title VII of the Civil Rights Act of 1964*. Available at https://www.eeoc.gov/laws/guidance/select-issues-assessing-adverse-impact-software-algorithms-and-artificial

Fan, J., Sun, T., Liu, J., Zhao, T., Zhang, B., Chen, Z., Glorioso, M., & Hack, E. (2023). How well can an AI chatbot infer personality? Examining psychometric properties of machine-inferred personality scores. *Journal of Applied Psychology, 108*(8), 1277–1299. https://doi.org/10.1037/apl0001082

Feffer, M. (2016). *Algorithms are changing the recruiter's role*. Available at https://www.shrm.org/resourcesandtools/hr-topics/talent-acquisition/pages/algorithms-changing-recruiters-role.aspx

Feffer, M., Hirzel, M., Hoffman, S. C., Kate, K., Ram, P., & Shinnar, A. (2022). An empirical study of modular bias mitigators and ensembles. *arXiv preprint arXiv:2202.00751*.

Feldman, T., & Peake, A. (2021). End-to-end bias mitigation: Removing gender bias in deep learning. *arXiv preprint arXiv:2104.02532*.

Fu, R., Liang, Y., & Zhang, P. (2021). Model mis-specification and algorithmic bias. *arXiv preprint arXiv:2105.15182*.

Garg, N., Schiebinger, L., Jurafsky, D., & Zou, J. (2018). Word embeddings quantify 100 years of gender and ethnic stereotypes. *Proceedings of the National Academy of Sciences*, *115*(16), E3635-E3644.

Glazko, K., Mohammed, Y., Kosa, B., Potluri, V., & Mankoff, J. (2024, June). Identifying and Improving Disability Bias in GPT-Based Resume Screening. In *The 2024 ACM Conference on Fairness, Accountability, and Transparency* (pp. 687-700). https://doi.org/10.1145/3630106.3658933

Hardt, M., Price, E., & Srebro, N. (2016). Equality of opportunity in supervised learning. *Advances in Neural Information Processing Systems*, *29*.

Harris, K. D., Murray, P., & Warren, E. (2018). *Letter to U.S. Equal Employment Opportunity Commission regarding risks of facial recognition technology*. Retrieved from https://www.scribd.com/document/388920670/SenHarris-EEOC-Facial-Recognition-2

Hellman, D. (2020). Measuring algorithmic fairness. *Virginia Law Review*, *106*(4), 811-866.

ALGORITHMIC BIAS MITIGATION 47what?,""What's new?," and "Where to next?" *Journal of Management*, *43*(6), 1693-1725.

Mehrabi, N., Morstatter, F., Saxena, N., Lerman, K., & Galstyan, A. (2021). A survey on bias and fairness in machine learning. *ACM Computing Surveys (CSUR)*, *54*(6), 1-35.

Minot, J. R., Cheney, N., Maier, M., Elbers, D. C., Danforth, C. M., & Dodds, P. S. (2022). Interpretable bias mitigation for textual data: Reducing genderization in patient notes while maintaining classification performance. *ACM Transactions on Computing for Healthcare*, *3*(4), Article 29. https://doi.org/10.1145/3524887

Miroshnikov, A., Kotsiopoulos, K., Franks, R., & Kannan, A. R. (2021). Model-agnostic bias mitigation methods with regressor distribution control for Wasserstein-based fairness metrics. *arXiv preprint arXiv:2111.11259*.

Mitchell, M., Wu, S., Zaldivar, A., Barnes, P., Vasserman, L., Hutchinson, B., Spitzer, E., Raji, I. D., & Gebru, T. (2019, January). Model cards for model reporting. In *Proceedings of the conference on fairness, accountability, and transparency* (pp. 220-229).

Morris, S. (2016). Statistical significance testing in adverse impact analysis. In S. M. Morris & E. M. Dunleavy (Eds.), *Adverse Impact Analysis: Understanding data, statistics, and risk* (pp. 91-111). Routledge.

Mujtaba, D. F., & Mahapatra, N. R. (2019, November). Ethical considerations in AI-based recruitment. In *2019 IEEE International Symposium on Technology and Society (ISTAS)* (pp. 1-7). IEEE.

Nye, C. D., & Sackett, P. R. (2017). New effect sizes for tests of categorical moderation and differential prediction. *Organizational Research Methods*, *20*(4), 639-664.

Obermeyer, Z., Powers, B., Vogeli, C., & Mullainathan, S. (2019). Dissecting racial bias in an algorithm used to manage the health of populations. *Science, 366*(6464), 447-453.

Oswald, F. L., Dunleavy, E. M., & Shaw, A. (2016). Measuring practical significance in adverse impact analysis. In S. M. Morris & E. M. Dunleavy (Eds.), *Adverse Impact Analysis: Understanding data, statistics, and risk* (pp. 112-132). Routledge.

Pfeifer Jr, C. M., & Sedlacek, W. E. (1971). The validity of academic predictors for black and white students at a predominantly white univertsity. *Journal of Educational Measurement*, *8*(4), 253-261. https://doi.org/10.1111/j.1745-3984.1971.tb00934.x

Pleiss, G., Raghavan, M., Wu, F., Kleinberg, J., & Weinberger, K. Q. (2017). On fairness and calibration. *Advances in Neural Information Processing Systems*, *30*.

Putka, D. J., Beatty, A. S., & Reeder, M. C. (2018). Modern prediction methods: New perspectives on a common problem. *Organizational Research Methods*, *21*(3), 689-732.

ALGORITHMIC BIAS MITIGATION    51Waller, M., Rodrigues, O., & Cocarascu, O. (2023). Bias Mitigation Methods for Binary Classification Decision-Making Systems: Survey and Recommendations. *arXiv preprint arXiv:2305.20020*.

Wainer, H. (1976). Estimating coefficients in linear models: It don't make no nevermind. *Psychological Bulletin*, *83*(2), 213-217.

Wainer, H. (2000). CATs: Whither and whence. *Psicologica*, 21(1), 121-133.

Wan, M., Zha, D., Liu, N., & Zou, N. (2021). Modeling techniques for machine learning fairness: A survey. *arXiv preprint arXiv:2111.03015*.

Xiang, F., Zhang, X., Cui, J., Carlin, M., & Song, Y. (2022, December). Algorithmic Bias in a Student Success Prediction Models: Two Case Studies. In *2022 IEEE International Conference on Teaching, Assessment and Learning for Engineering (TALE)* (pp. 310-315). IEEE.

Yarkoni, T., & Westfall, J. (2017). Choosing prediction over explanation in psychology: Lessons from machine learning. *Perspectives on Psychological Science*, *12*(6), 1100-1122.

Zehlike, M., Ke Yang, & Stoyanovich, J. (2023). Fairness in Ranking, Part II: Learning-to-Rank and Recommender Systems. *ACM Computing Surveys,* *55*(6), 1–41. https://doi.org/10.1145/3533380

Zemel, R., Wu, Y., Swersky, K., Pitassi, T., & Dwork, C. (2013, May). Learning fair representations. In *International Conference on Machine Learning* (pp. 325-333). PMLR.

Zhang, N., Wang, M., Xu, H., Koenig, N., Hickman, L., Kuruzovich, J., Ng, V., Arhin, K., Wilson, D., Song, Q. C., Tang, C., Alexander III, L., & Kim, Y. (2023). Reducing subgroup differences in personnel selection through the application of machine learning. *Personnel Psychology*, *76*(4), 1125-1159.

**Tables**

**Table 1**

*Key Terms and Definitions Relevant to Algorithmic Bias Mitigation*

| Term | Definition |
|---|---|
| ML Algorithm | Procedures that can be run on data (i.e., "trained") to identify patterns and create a ML model. These are broadly categorized as either supervised ML, unsupervised ML, or reinforcement learning techniques. |
| ML Model | The outputs of ML algorithms which have been trained on data, consisting of parameters and a mathematical function that determine which output is generated from a given input. |
| Supervised ML | Statistical models developed by relating independent (predictor) variables to an outcome (dependent) variable. Examples include ordinary least squares regression, random forest, and neural networks. |
| Unsupervised ML | Statistical models developed by examining how independent (predictor) variables covary, and no dependent variable is used. Examples include clustering, factor analysis, and latent profile analysis. |
| Reinforcement learning | A type of machine learning where an "agent" iteratively interacts with the environment, receives feedback (i.e., rewards or penalties), and learns to adjust its behavior to maximize rewards. |
| Psychometric Bias | Systematic error in test scores that differentially affects test takers as a function of group membership (SIOP, 2018). |
| Predictive Bias | Systematic prediction errors for individuals from a specific subgroup. |
| Measurement Bias | Systematically higher/lower scores for individuals from one or more subgroups. In ML, can also mean systematic measurement errors (that may not affect mean scores; Tay et al., 2022). |
| Fairness | A subjective social concept that has multiple meanings, depending on what one perceives to be fair (SIOP, 2018). Fairness notions include equity, equality, and need, as well as equal group outcomes, a lack of bias, equitable treatment, and comparable access to constructs (SIOP, 2018). "Algorithmic bias" is frequently used in a way that encompasses fairness *and* bias. |

ALGORITHMIC BIAS MITIGATION                                                53| | |
|---|---|
| Adverse Impact | Subgroup differences in selection ratios. Selection ratios that are statistically significantly different from each other, or where one is four-fifths of the other, may constitute *prima facie* evidence of discrimination in the United States. |
| Fairness Operationalizations | Specific, operational definitions of algorithmic bias, mostly originating from computer and data science. These include characteristics of the predictors, trained models, measures of the ML model scores' psychometric bias, and adverse impact. |
| Algorithmic Bias Mitigation | ML methods and design decisions that aim to reduce bias and/or adverse impact, usually by targeting a fairness operationalization. These can occur during data collection or curation (i.e., preprocessing), during model training (i.e., inprocessing), or after model predictions are made (i.e., postprocessing) |
| Ground Truth | The observed outcome (or dependent, $y$) variable used during model training. |
| Explainable AI (XAI) | Methods for explaining ML model functioning or outputs in ways that humans understand. |

*Note*. ML = machine learning.

ALGORITHMIC BIAS MITIGATION 54**Table 2**

*Four Stages of the ML Model Life Cycle*

| Step | Description | Representative Considerations |
|---|---|---|
| Generating Training Data | Collecting or selecting (from archival data) a dataset for training ML models. | ● Is the training sample representative of the people likely to be encountered during deployment?<br>● Are the predictor measures fair and unbiased?<br>● Is the dependent variable measure fair and unbiased?<br>● Are group differences on predictors or the dependent variable unexpectedly large? |
| Training the Model | Fitting statistical models to the training data. | ● Are the predictors conceptually related to the dependent variable?<br>● Do the models use demographic information (either directly or indirectly) to score individuals?<br>● Can the model and its scores be explained?<br>● What optimization function(s) was used? |
| Testing the Model | Applying trained ML models to other data (i.e., a separate sample or derived via splitting the training data) and evaluating the model(s) predictions' psychometric properties (i.e., cross-validation). | ● Is the test sample representative of the people likely to be encountered during deployment?<br>● Was the test sample collected concurrent to the training data or temporally lagged, as will occur during deployment?<br>● How do group differences and adverse impact compare to the ground truth dependent variable?<br>● Is psychometric (measurement or predictive) bias present? |
| Deploying the Model | Using a trained ML model to assess people in high-stakes settings. | ● Are the psychometric properties of the ML model scores consistent during deployment and over time?<br>● Are the ML models used consistently in the decision-making process?<br>● Are people aware they are being evaluated by the ML model, and is the model and its characteristics explained to them? |

**Table 3**

*Sources of Algorithmic Bias*

| Source | Description |
| --- | --- |
| Aggregation Bias | Occurs when one ML model is applied to multiple groups, even though the true predictor-outcome relationships differ across groups. For example, a model that uses organizational commitment to predict turnover may exhibit aggregation bias because organizational commitment may be less related to turnover among women than men (Russ & McNeilly, 1995). |
| Emergent Bias | Occurs when the true nature of predictor-outcome relationships change after an ML model has been deployed. This can occur because of changing populations and culture values, or because of updated knowledge (e.g., a subset of users gains access to resources that help them exploit the ML scoring system). |
| Historical Bias | Occurs when what truly is and has been in the world leads the model to accurately reflect the world but to inflict harm on a subgroup. For example, word embeddings are trained on human data that reflect human biases which causes differential associations between group labels and other concepts (Garg et al., 2018). |
| Human Bias | Occurs when input or ground truth variables rely on human ratings that were biased by construct-irrelevant factors. For example, interviewer evaluations may be affected by how attractive interviewees are (Hosoda et al., 2003). |
| Learning Bias | Occurs when the ML model exacerbates score differences across groups. This may occur, for example, when model predictions accurately recover group means but exhibit smaller variances than observed ground truth scores (e.g., Fan et al., 2023) |
| Measurement Bias | Occurs when (a) predictor or ground truth variables exhibit psychometric measurement bias, (b) when irrelevant variables are included in the model, (c) when relevant variables are omitted from the model, and (d) may occur when proxy variables are used instead of the actual variable. |
| Representation Bias | Occurs when some portion of the population is underrepresented in training data. For example, an employment interview scoring system may be less accurate for minority groups, even when they are represented proportionally to their presence in the population (Zhang et al., 2023). |

*Note.* Sources include Suresh and Guttag (2021), Booth et al. (2021), and Köchling and Wehner (2020).

**Table 4**

*Fairness Operationalizations Across the Stages of the ML Model Life Cycle*

| Step | Fairness Term | Description |
|---|---|---|
| Generating Training Data: Characteristics of the training data | Equivalence of computed features | Predictor variables should not carry undue amounts of information about group membership, either because of measurement bias, historical bias, or otherwise. |
| | Adverse impact | Large ground truth group differences (i.e., adverse impact; Uniform Guidelines, 1978) may encode group differences in the ML model. |
| | Training sample representativeness | When demographic groups are disproportionately represented in training data, the ML model may overrepresent predictor-outcome patterns for the majority group. |
| | Potential bias | Extent that group membership can be predicted from the training data. |
| Training the Model: Predictor choice and causal influence on ML outcomes | Causal fairness | Does group membership cause different ML model outcomes? |
| | Fairness through unawareness | ML models are fair that do not include protected demographics as predictors. |
| | Disparate treatment | The same model must be applied to all, and protected demographics cannot be used as predictors in ML models. |
| | Counterfactual fairness; no unresolved discrimination | ML models are fair when their outcomes/decisions do not depend on one or more causal descendants of protected demographics. |
| | Fair inference | Predictor variables that are causal dependents of protected demographics must be causally related to the outcome variable to be justifiable. |
| Testing the Model: Group- and individual-level consistency | Group fairness | Does the ML model exhibit similar score distributions (independence) or validity (sufficiency and separation) across groups? |
| | Adverse impact | ML model scores should not exhibit large group differences, especially if they would violate the four-fifths rule (Uniform Guidelines, 1978). |
| | Conditional statistical parity | ML model scores should not exhibit large group differences conditional on one or more legitimate variables. |
| | Equal accuracy | ML model scores should be similarly accurate and valid across demographic groups; a lack of measurement and predictive bias. Includes group comparisons of elements of confusion matrices. |
| | Bias and differential functioning | ML model scores should not exacerbate group differences compared to the ground truth scores they were trained to replicate, and they should be equally predictive of outcomes across groups (i.e., no predictive bias). |
| | Individual fairness/Consistency | Also known as fairness through awareness: Similar individuals should receive similar ML scores regardless of demographics. Examine score consistency with near neighbors. |
| Deploying the Model | Causal fairness | See above. |
| | Group fairness | See above. |
| | Individual fairness/Consistency | See above. |

**Table 5**

*Bias Mitigation Methods in the Generating the Training Data Stage*

| Bias Mitigation Method | Fairness Operationalizations Addressed | Legality |
|---|---|---|
| Balancing representation<br>  Sampling<br>  Resampling/Oversampling<br>  Undersampling/Downsampling<br>  Reweighing | Training sample representativeness | • Adjusting training data representation is legal. |
| Removing demographic information from the predictor variables<br>  Removing predictors<br>  Transforming predictors (group-specific)<br>    Learning fair representations<br>    Residualize on demographics<br>    Adversarially fair representations<br>    Disparate impact remover<br>    FairGAN | Fairness through unawareness; equivalence of computed features; adverse impact; potential bias | • Not including demographic information as a predictor is required by statutes outlawing disparate impact.<br>• Removing predictors that contribute to group differences could legally help reduce adverse impact.<br>• Transforming predictors with group-specific methods is illegal if also applied during deployment, because different transformations are applied to an individual's data depending on their group membership, thus constituting disparate treatment and/or subgroup norming (Rottman et al., 2023). |
| Removing demographic information from the outcome variables<br>  Sampling<br>  Resampling/Oversampling<br>  Undersampling/Downsampling<br>  Reweighing<br>  Massaging ground truth labels | Adverse impact | • Reducing group disparities in the training data ground truth could help ameliorate adverse impact, and the legality of training on data with large adverse impact could be challenged.<br>• Massaging (i.e., changing) group truth labels, while likely legal, may harm validity and utility. |

*Note.* Under each category of methods, representative approaches are listed, but the lists are not comprehensive. Additional details about specific methods and the frequency they appeared in the literature review are provided additional online material.

ALGORITHMIC BIAS MITIGATION                                    58**Table 6**

*Bias Mitigation Methods in the Training the Model Stage*

| Bias Mitigation Method | Fairness Operationalizations Addressed | Legality |
|---|---|---|
| Multi-objective optimization<br>    Prejudice remover regularizer<br>    Adversarial debiasing<br>    Pareto-optimization<br>    Meta fair classifier | Depends on the optimization function used but can address adverse impact, equal accuracy, and consistency. | • Adjusting test design to maximize diversity and validity is legal. |
| Separate models or including demographic information as a predictor variable<br>    Domain discriminative training<br>    Transfer learning | Adverse impact; equal accuracy; differential functioning. | • Constitutes disparate treatment and/or subgroup norming so is likely illegal in the United States and Europe. |

*Note.* Under each category of methods, representative approaches are listed, but the lists are not comprehensive. Additional details about specific methods and the frequency they appeared in the literature review are provided in the additional online material.

**Table 7**

*Bias Mitigation Methods in the Testing the Model Stage*

| Bias Mitigation Method | Fairness Operationalizations Addressed | Legality |
|---|---|---|
| Group-specific postprocessing | Adverse impact; equal accuracy; differential functioning. | • These methods apply different cut scores or transformations depending on an individual's demographics and are, thus, illegal because they constitute disparate treatment and/or subgroup norming. |
| Group-agnostic postprocessing | Adverse impact | • Transformations that alter everyone's ML scores in the same way are legal (but may not have much effect on adverse impact). |

*Note.* Under each category of methods, representative approaches are listed, but the lists are not comprehensive. Additional details about specific methods and the frequency they appeared in the literature review are provided in the additional online material.

**Table 8**

*Bias Mitigation Methods in the Deploying the Model Stage*

| Bias Mitigation Method | Fairness Operationalizations Addressed | Legality |
|---|---|---|
| Score Banding | Adverse impact | Legal to use score banding but not to make decisions within bands based on protected characteristics. |

Table 9

*Best Practice Recommendations for Developing ML Models for High-Stakes Assessment*

| **Throughout the Process** |
|---|
| • Document all decisions made and code used. |
| **Generating the Training Data** |
| • Collect diverse training data that is representative of the intended population.<br>• Minimize adverse impact in training data (e.g., through purposeful sampling or oversampling).<br>• Examine predictors and ground truth variables for measurement and predictive bias<br>• Critically evaluate predictors with sizeable group differences. |
| **Training the Model** |
| • Do not use demography as a predictor, to make group-specific transformations, to slot people into different evaluation procedures, or to adjust their scores.<br>• Consider the meaningfulness and content relevance of predictors. For constructs, use work analysis. For behavioral and trace data, consider the relevance of the data collection context to the work that will be performed.<br>• During model selection, hyperparameter tuning, and with multi-objective optimization, use demography to make choices that optimize for both validity (i.e., accuracy/convergence) and diversity.<br>• Additional model complexity that reduces model interpretability should be justified by validity gains. |
| **Testing the Model** |
| • Evaluate multiple aspects of the validity for the ML model and its scores.<br>• Check ML model scores' adverse impact, equal accuracy, and differential functioning.<br>• Consider examining the ML model scores' psychometric properties in a lagged test sample prior to deployment. |
| **Deploying the Model** |
| • Continuously check the ML model scores' validity, adverse impact, and for equal accuracy and differential functioning.<br>• Monitor for demographic, occupation, industry, and qualification-level differences between training/testing data and data encountered during deployment.<br>• If substantial changes in psychometric properties occur, consider generating new training data and/or retraining the model. |

# Figures

**Figure 1**

*Four-stage model integrating fairness concepts, United States legal requirements, and algorithmic bias mitigation methods*

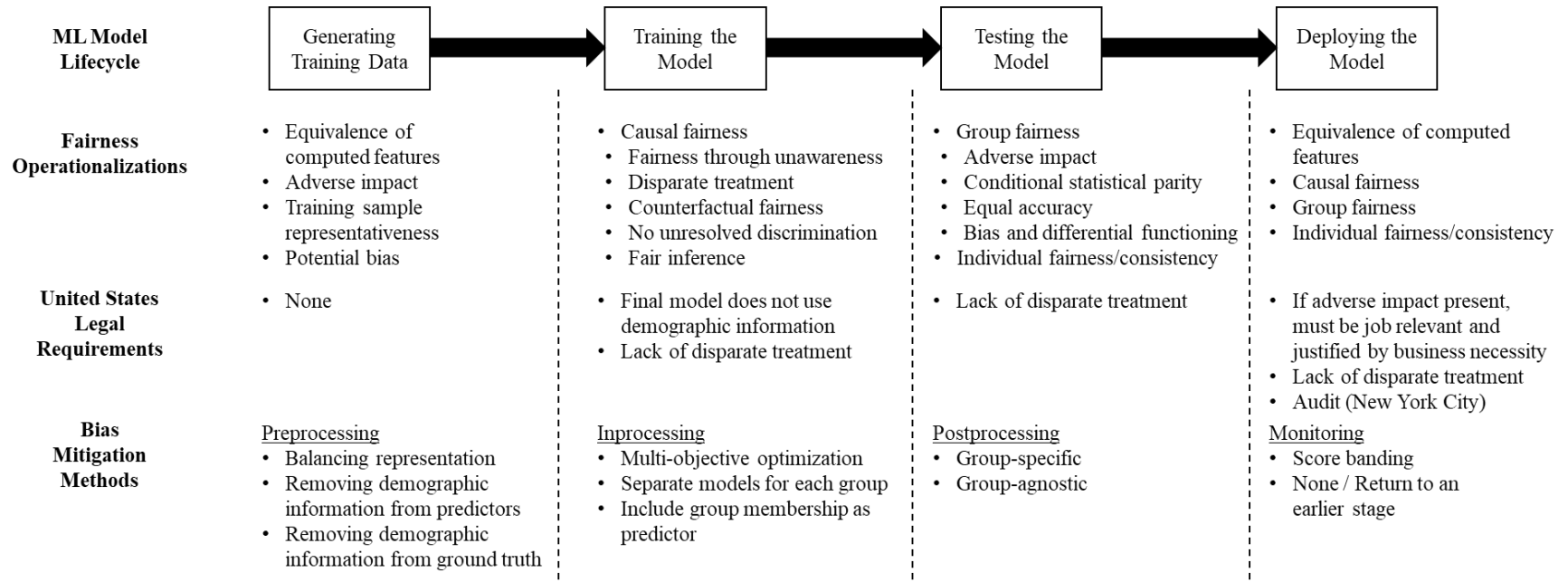

*Note.* Although the bias mitigation methods listed under "Testing the Model" could also be listed under "Deploying the Model," we do not list them there because they all explicitly use demographic information to adjust ML model outputs and, therefore, are not legal to use for personnel selection in the United States or Europe because they constitute disparate treatment and/or subgroup norming.

**Figure 2**

*Linkages between concepts and perspectives on bias and fairness*

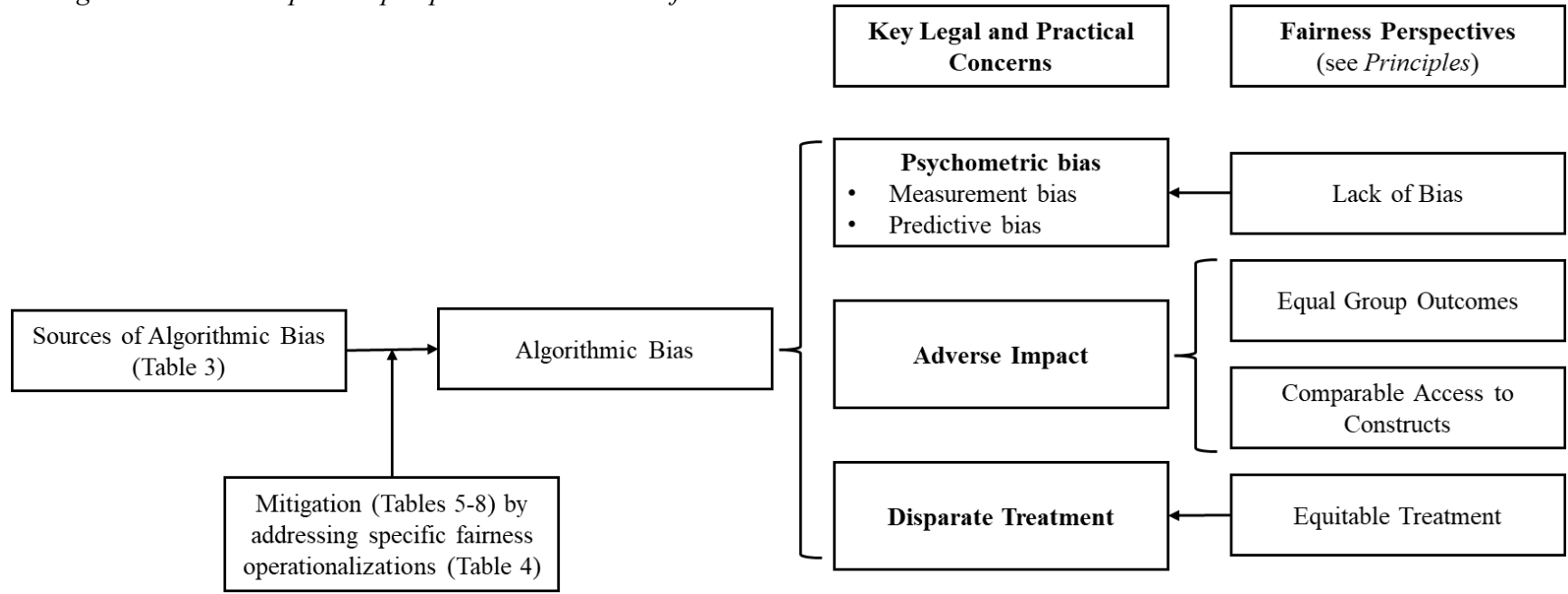

Figure 3

*PRISMA systematic literature search flow diagram*

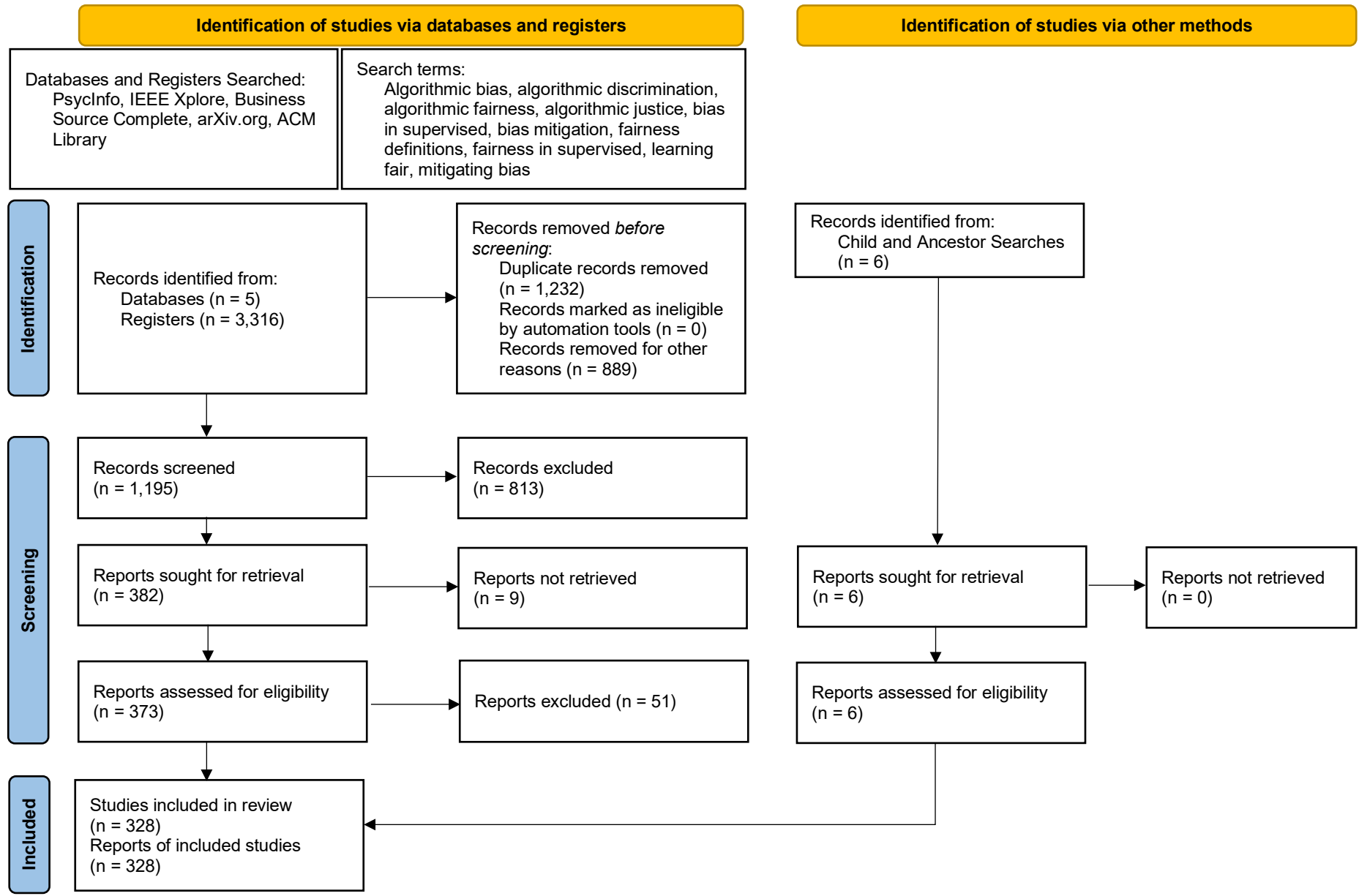